%% file: main.tex
\documentclass[11pt,a4paper]{article}
\usepackage{times,latexsym}
\usepackage{url}
\usepackage[T1]{fontenc}

\usepackage[acceptedWithA]{tacl2021v1}

\usepackage{xspace,mfirstuc,tabulary}
\usepackage{hyperref}       %
\usepackage{booktabs}       %
\usepackage{amsfonts}       %
\usepackage{nicefrac}       %
\usepackage{microtype}      %
\usepackage{xcolor}         %
\usepackage{graphicx}
\usepackage{amsmath}
\usepackage{subcaption,multirow,multicol}
\usepackage{caption} 
\usepackage{longtable}
\usepackage{adjustbox}
\usepackage[toc,page]{appendix}

\newif\iftaclinstructions
\taclinstructionsfalse %
\iftaclinstructions
\renewcommand{\confidential}{}
\renewcommand{\anonsubtext}{(No author info supplied here, for consistency with
TACL-submission anonymization requirements)}
\newcommand{\instr}
\fi

\iftaclpubformat %

\else

\fi

\definecolor{darkgreen}{HTML}{6aa84f}
\newcommand{\ku}{$^\dagger$}
\newcommand{\mbzuai}{$^\ddagger$}

\title{Do Vision and Language Models Share Concepts? \\ A Vector Space Alignment Study}

\author{Jiaang Li\ku \ \ Yova Kementchedjhieva\mbzuai \ \ Constanza Fierro\ku \ \ Anders Søgaard\ku\\ \\
{\ku} University of Copenhagen \\
{\mbzuai} Mohamed bin Zayed University of Artificial Intelligence \\
\texttt{\{jili,c.fierro,soegaard\}@di.ku.dk,  yova.kementchedjhieva@mbzuai.ac.ae}
}

\date{}

\begin{document}
\maketitle
\begin{abstract}
Large-scale pretrained language models (LMs) are said to ``lack the ability to connect utterances to the world''~\cite{bender-koller-2020-climbing}, because they do not have ``mental models of the world''~\cite{doi:10.1073/pnas.2215907120}. If so, one would expect LM representations to be unrelated to representations induced by vision models.
We present an empirical evaluation across four families of LMs (BERT, GPT-2, OPT and LLaMA-2) and three vision model architectures (ResNet, SegFormer, and MAE). Our experiments show that LMs partially converge towards representations isomorphic to those of vision models, subject to dispersion, polysemy and frequency. This has important implications for {\em both} multi-modal processing and the LM understanding debate~\cite{doi:10.1073/pnas.2215907120}.\footnote{Code and dataset: \url{https://github.com/jiaangli/VLCA}.}
\end{abstract}

\input{sections/01_intro}
\input{sections/02_relate}
\input{sections/03_method}
\input{sections/04_exp}
\input{sections/05_res}

\input{sections/06_analysis}

\input{sections/07_discussion}

\input{sections/08_conclusion}

\section*{Acknowledgments}
We thank the reviewers and action editors for their invaluable feedback. Special thanks to Serge Belongie and Vésteinn Snæbjarnarson for helpful discussions. Jiaang Li is supported by Carlsberg Research Foundation (grant: CF221432) and the Pioneer Centre for AI, DNRF grant number P1.

\bibliography{custom}
\bibliographystyle{acl_natbib}

\input{sections/09_appendix}

\end{document}

%% file: sections/01_intro.tex
\section{Introduction}
The debate around whether LMs can be said to understand is often portrayed as a back-and-forth between two opposing sides~\cite{doi:10.1073/pnas.2215907120}, but in reality, there are many positions. Some researchers have argued that LMs are `all syntax, no semantics', i.e., that they learn form, but not meaning~\cite{searle80minds,bender-koller-2020-climbing,marcus2023sentence}.\footnote{The idea that computers are `all syntax, no semantics' can be traced back to German 17th century philosopher Leibniz's Mill Argument~\cite{10.2307/27903609}. The Mill Argument states that mental states cannot be reduced to physical states, so if the capacity to understand language requires mental states, this capacity cannot be instantiated, merely imitated, by machines. In 1980, Searle introduced an even more popular argument against the possibility of LM understanding, in the form of the so-called Chinese Room thought experiment~\cite{searle80minds}. The Chinese Room presents an interlocutor with no prior knowledge of a foreign language, who receives text messages in this language and follows a rule book to reply to the messages. The interlocutor is Searle’s caricature of artificial intelligence, and is obviously, Searle claims, not endowed with meaning or understanding, but merely symbol manipulation.} Others have argued that LMs have inferential semantics, but not referential semantics~\cite{Rapaport2002-RAPHCS,sahlgren2021singleton,piantadosi2022meaning},\footnote{See \citet{marconi1997lexical} for this distinction.} whereas some have posited that a form of externalist referential semantics is possible, at least for chatbots engaged in direct conversation~\cite{Cappelen2021-CAPMAI,Butlin2021-BUTSOC-2,mollo2023vector,mandelkern2023language}. Most researchers agree, however, that LMs ``lack the ability to connect utterances to the world''~\cite{bender-koller-2020-climbing}, because they do not have ``mental models of the world''~\cite{doi:10.1073/pnas.2215907120}. %

This study provides evidence to the contrary: Language models and computer vision models (VMs) are trained on independent data sources (at least for unsupervised computer vision models). The only common source of bias is the world. If LMs and VMs exhibit similarities, it must be because they both model the world. We examine the representations learned by different LMs and VMs by measuring how similar their geometries are. We consistently find that the better the LMs are, the more they induce representations similar to those induced by computer vision models. The similarity between the two spaces is such that from a very small set of parallel examples we are able to linearly project VMs representations to the language space and retrieve highly accurate captions, as shown by the examples in Figure~\ref{fig:res-examples}.

\paragraph{Contributions.} We present a series of evaluations of the vector spaces induced by three families of VMs
and four families of LMs, 
i.e., a total of fourteen VMs and fourteen LMs. We show that within each family, the larger the LMs, the more their vector spaces become structurally similar to those of computer vision models. This enables retrieval of language representations of images (referential semantics) with minimal supervision. Retrieval precision depends on dispersion of image and language, polysemy, and frequency, but consistently improves with language model size. We discuss the implications of the finding that language and computer vision models learn representations with similar geometries.

%% file: sections/02_relate.tex
\section{Related Work}

\paragraph{Inspiration from cognitive science.}
Computational modeling is a cornerstone of cognitive science in the pursuit for a better understanding of how representations in the brain come about. As such, the field has shown a growing interest in computational representations induced with self-supervised learning~\cite{NEURIPS2020_7183145a,Halvagal2022.03.17.484712}. 
Cognitive scientists have also noted how the objectives of supervised language and vision models bear resemblances to predictive processing~\cite{Schrimpf_brain_score_2018, Goldstein2020.12.02.403477, Caucheteux22LongrangeAH,li2023structural} (but see \citet{antonello_why} for a critical discussion of such work). 

Studies have looked at the alignability of neural language representations and human brain activations, with more promising results as language models grow better at modeling language~\cite{10.1162/nol_a_00003,Schrimpf2020.06.26.174482}. In these studies, the partial alignability of brain and model representations is interpreted as evidence that brain and models might process language in the same way~\cite{CaucheteuxC2022BrainsAA}.

\paragraph{Cross-modal alignment.} The idea of cross-modal retrieval is not new~\cite{lazaridou-etal-2014-wampimuk}, but previously it has mostly been studied with practical considerations in mind. Recently, \citet{merullo2023linearly} showed that language representations in LMs are \textit{functionally} similar to image representations in VMs, in that a linear transformation applied to an image representation can be used to prompt a language model into producing a relevant caption. We dial back from function and study whether the concept representations converge toward structural similarity (isomorphism). The key question we address is whether despite the lack of explicit grounding, the representations learned by large pretrained language models structurally resemble properties of the physical world as captured by vision models. More related to our work, \citet{huh2024platonic} proposes a similar hypothesis, although studying it from a different perspective, and our findings corroborate theirs.

\begin{figure}
    \centering
    \includegraphics[width=1\linewidth]{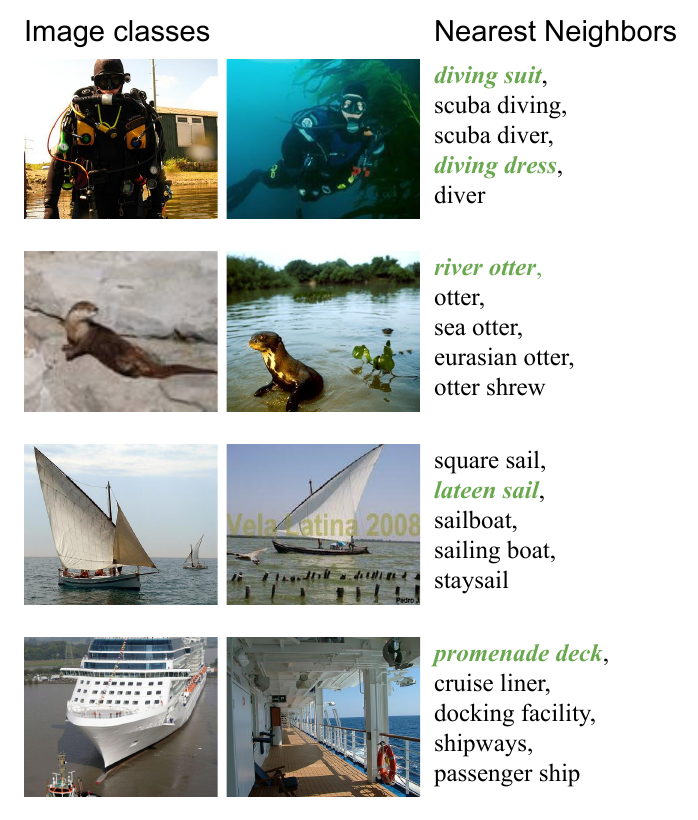}
    \caption{Mapping from MAE$_{\text{Huge}}$ (images) to OPT$_{\text{30B}}$ (text). Gold labels are in \textcolor{darkgreen}{green}. 
    }
    \label{fig:res-examples}
\end{figure}

%% file: sections/03_method.tex
\section{Methodology\label{section:methodology}}
\begin{figure*}
    \centering
    \includegraphics[width=1\textwidth]{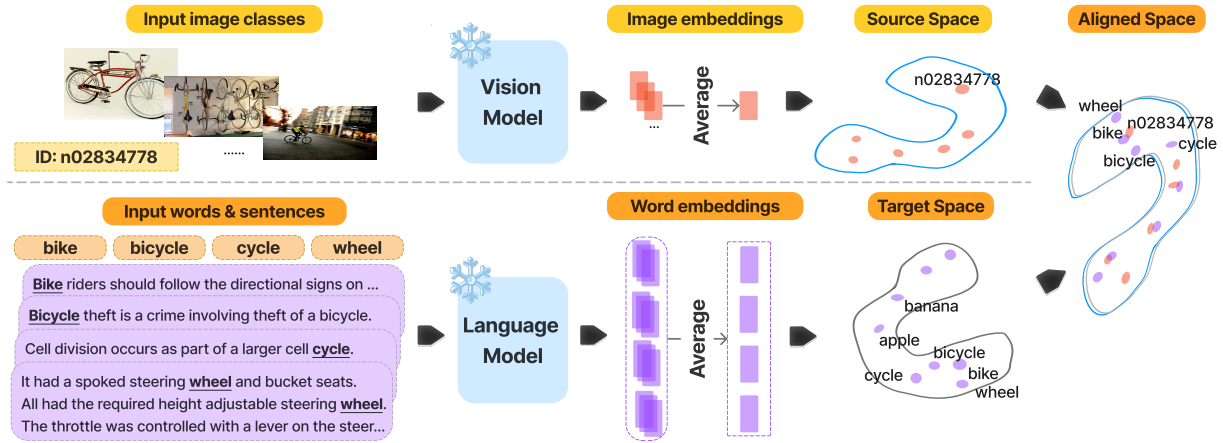}
    \caption{Experiments stages: During our experiments, words, sentences, and images are selected from the aliases list (wordlist and ImageNet-21K aliases), Wikipedia and ImageNet-21K, respectively. The source and target spaces are constructed utilizing image and word embeddings which are extracted by specialized vision and language models.}
    \label{fig:pipeline}
\end{figure*}

Our primary objective is to compare the representations derived from VMs and LMs and assess their alignability, i.e. the extent to which LMs converge toward VMs' geometries. %
In the following sections, we introduce the procedures for obtaining the representations and aligning them, with an illustration of our methodology provided in Figure~\ref{fig:pipeline}.

\paragraph{Vision models.}
We include fourteen VMs in our experiments, representing three model families: SegFormer~\cite{NEURIPS2021_64f1f27b}, MAE~\cite{he2022masked}, and ResNet~\cite{he2016residual}. 
For all three types of VMs, we only employ the encoder component as a visual feature extractor.\footnote{We ran experiments with CLIP~\cite{radford2021learning}, but report on these separately, since CLIP does not meet the criteria of our study, being trained on a mixture of text and images. CLIP results are presented in Appendix~\ref{sec:appendix-clip-results}.} 

SegFormer models consist of a Transformer-based encoder and a light-weight feed-forward decoder. They are pretrained on object classification data and finetuned on scene parsing data for scene segmentation and object classification. We hypothesize that the reasoning necessary to perform segmentation in context promotes representations that are more similar to those of LMs, which also operate in a discrete space (a vocabulary). The SegFormer models we use are pretrained with ImageNet-1K~\cite{ILSVRC15} and finetuned with ADE20K~\cite{BoleiZhou2017ScenePT}. 

MAE models relies on a Transformer-based encoder-decoder architecture, with the Vision-Transformer (ViT)~\cite{DBLP:conf/iclr/DosovitskiyB0WZ21} as the encoder backbone. MAE models are trained to reconstruct masked patches in images, i.e., a fully unsupervised training objective, similar to masked language modeling. The encoder takes as input the unmasked image patches, while a lightweight decoder reconstructs the original image from the latent representation of unmasked patches interleaved with mask tokens. The MAE models we use are pretrained on ImageNet-1K. 

ResNet models for object classification consist of a bottleneck convolutional neural network with residual blocks as an encoder, with a classification head. They are pretrained on the ImageNet-1K. 

\paragraph{Language models.} We include fourteen Transformer-based LMs in our experiments, representing four model families: BERT~\citep{devlin-etal-2019-bert}, GPT-2~\citep{radford2019language}, OPT~\citep{zhang2022opt} and LLaMA-2~\citep{touvron2023llama}. 
We use six different sizes of BERT (all uncased): BERT$_{\text{Base}}$ and BERT$_{\text{Large}}$, which are pretrained on the BooksCorpus~\cite{Zhu2015AligningBA} and English Wikipedia~\cite{wikidump}, and four smaller BERT sizes, distilled from BERT$_{\text{Large}}$~\cite{turc2019}. GPT-2, an auto-regressive decoder-only LM, comes in three sizes,
pretrained on the WebText dataset~\cite{radford2019language}. OPT also %
comes in three sizes, pretrained on the union of five datasets~\citep{zhang2022opt}. %
LLaMA-2 was pretrained on two trillion tokens. %

\paragraph{Vision representations.}
The visual representation of a concept is obtained by embedding
the images available for the concept with a given VM encoder and then averaging these representations. 
When applying SegFormer, we average the patches' representations from the last hidden state as the basis for every image, whereas we use the penultimate hidden state for MAE models.\footnote{We also experimented with utilizing the representations from the last hidden state; however, the results were not as promising as those obtained from the penultimate hidden state. \citet{caron2021emerging} demonstrate the penultimate-layer features in ViTs trained with DINO exhibit strong correlations with saliency information in the visual input, such as object boundaries and so on.} ResNet models generate a single vector per input image from the average pooling layer.

\paragraph{Language representations.}
The LMs included here were trained on text segments, so applying them to words in isolation could result in unpredictable behavior. We therefore represent %
words by embedding English Wikipedia sentences, using the token representations that form the concept, decontextualizing these representations by averaging across different sentences~\cite{abdou-etal-2021-language}. In the case of masked language models, we employ an averaging approach on the token representations forming the concept; otherwise, we choose for the last token within the concept~\cite{zou2023representation}. %

\paragraph{Linear projection.} Since we are interested in the extent to which vision and language representations are isomorphic, we focus on linear projections.\footnote{For work on non-linear projection between representation spaces, see \citet{nakashole-2018-norma,zhao-gilman-2020-non,glavas-vulic-2020-non}.} Following \citet{Conneau2018}, we use Procrustes analysis~\cite{schonemann1966procrustes} to align the representations of VMs to those of LMs, given a bimodal dictionary (\S~\ref{subsection:bimodal-dictionaries}). %
Given the VM matrix $A$ (i.e., the visual representations of concepts) and the LM matrix $B$ (i.e. the language representation of the concepts) we use Procrustes analysis to find the orthogonal matrix $\Omega$ that most closely maps source space $A$ onto the target space $B$. Given the constrain of orthogonality the optimization $\Omega = \min_{R}\|RA-B\|_{F} , \text{s.t.} \ R^{T}R=I$ has the closed form solution $\Omega=UV^{T}, U\Sigma V=\text{SVD}(BA^T)$, where SVD stands for singular value decomposition. We induce the alignment from a small set of dictionary pairs, evaluating it on held-out data (\S~\ref{subsection:evaluation}). Given the necessity for both the source and target space to have the same dimensionality, we employ principal component analysis (PCA) to reduce the dimensionality of the larger space in cases of a mismatch.%
\footnote{The variance is retained for most models after dimensionality reduction, except for a few cases where there is some loss of information. The cumulative of explained variance ratios for different models are presented in Table~\ref{tab:pca-var-ratio}.}

%% file: sections/04_exp.tex
\section{Experimental Setup}

In this section, we discuss details around bimodal dictionary compilation (\S~\ref{subsection:bimodal-dictionaries}), evaluation metrics, as well as our baselines (\S~\ref{subsection:evaluation}).

\subsection{Bimodal Dictionary Compilation\label{subsection:bimodal-dictionaries}}
We build bimodal dictionaries of image-text pairs based on the ImageNet-\textbf{21K} dataset~\cite{ILSVRC15} and the CLDI (cross-lingual dictionary induction) dataset~\cite{hartmann-sogaard-2018-limitations}. 
In ImageNet, a concept class has a unique ID and is represented by multiple images and one or more names (which we refer to as {\em aliases}), many of which are multi-word expressions. 
We filter the data from ImageNet-21K: keeping classes with over 100 images available, aliases that appear at least five times in Wikipedia, and classes with at least one alias. %
As a result, 11,338 classes and 13,460 aliases
meet the criteria. %
We further filter aliases that are shared by two different class IDs, and aliases for which their hyponyms are already in the aliases set.\footnote{We obtain the aliases hypernyms and hyponyms from the Princeton WordNet~\cite{fellbaum2010wordnet}.}
To avoid {\em any} form of bias, given that the VMs we experiment with have been pretrained on ImageNet-1K, we report results on ImageNet-21K excluding the concepts in ImageNet-1K (Exclude-1K). %

One important limitation of the Exclude-1K bimodal dictionary is that all concepts are nouns. 
Therefore, to investigate how our results generalize to other parts of speech (POS),
we also use the English subset of CLDI dataset (EN-CLDI),
which contains images paired with verbs and adjectives. Each word within this set is unique and paired with at least 22 images. Final statistics of the processed datasets are reported in Table~\ref{tab:bimoda_dictionaries_details}. 

The pairs in these bimodal dictionaries are split 70-30 for training and testing based on the class IDs to avoid train-test leakage.\footnote{In the EN-CLDI set, we simply use words to mitigate the risk of train-test leakage.} We compute five such splits at random and report averaged results. See \S~\ref{analysis:train-ratio} for the impact of training set size variations.

\begin{table}
\centering
\resizebox{1\linewidth}{!}{
\begin{tabular}{l|rrr}
\toprule
Set & \ Num. of classes & Num. of aliases & Num. of pairs\\
\midrule 
Only-1K & 491 & 655 & 655 \\
Exclude-1K & 5,942 & 7,194 & 7,194 \\
\midrule
EN-CLDI & 1,690 & 1,690 & 1,690 \\
\bottomrule 
\end{tabular}}
\caption{\label{tab:bimoda_dictionaries_details} Statistics of the bimodal dictionaries.}
\end{table}

\subsection{Evaluation\label{subsection:evaluation}}

We induce a linear mapping $\Omega$ based on training image-text pairs sampled from $A$ and $B$, respectively. We then evaluate how close $A\Omega$ is to $B$ by computing retrieval precision on held-out image-text pairs. 
To make the retrieval task as challenging as possible, the target space $B$ is expanded with 65,599 words from \href{https://www.npmjs.com/package/wordlist-english}{an English wordlist} in addition to 13,460 aliases,
resulting in a total of 79,059 aliases in the final target space. 

\paragraph{Metrics.} We evaluate alignment in terms of precision-at-$k$ (P@$k$), a well-established metric employed in the evaluation of multilingual word embeddings~\cite{Conneau2018}, with $k\in\{1,10,100\}$.\footnote{For example, we could use the mapping of the image of an apple into the word `apple', and the mapping of the image of a banana into the word `banana', as training pairs to induce a mapping $\Omega$. If $\Omega$ then maps the image of a lemon onto the word `lemon' as its nearest neighbor, we say that the precision-at-one for this mapping is 100\%. If two target aliases were listed in the bimodal dictionary for the source image, mapping the image onto either of them would result in P@$1=100\%$.} Note that this performance metric is much more conservative than other metrics used for similar problems, including pairwise matching accuracy, percentile rank, and Pearson correlation~\cite{minnema-herbelot-2019-brain}. Pairwise matching accuracy and percentile rank have random baseline scores of 0.5, and they converge in the limit. If $a$ has a percentile rank of $p$ in a list $\mathcal{A}$, it will be higher than a random member of $\mathcal{A}$ $p$ percent of the time. Pearson correlation is monotonically increasing with pairwise matching accuracy, but P@$k$ scores are more conservative than any of them for reasonably small values of $k$. In our case, our target space is 79,059 words, so it is possible to have P@100 values of 0.0 and yet still have near-perfect pairwise matching accuracy, percentile rank, and Pearson correlation scores. P@$k$ scores also have the advantage that they are intuitive and practically relevant, e.g., for decoding.

\begin{table}
\centering
\resizebox{1\linewidth}{!}{
\begin{tabular}{l|rrr}
\toprule
Baseline & P@1 & P@10 & P@100\\
\midrule 
Random retrieval & 0.0015 & 0.0153 & 0.1531 \\
Length-frequency alignment & 0.0032 & 0.0127 & 0.6053 \\
Non-isomorphic alignment & 0.0000 & 0.0121 & 0.1105\\
\bottomrule 
\end{tabular}}
\caption{\label{tab:baseline-details} Alignment results for our baselines. All the Precision@$k$ scores are reported in percentage.}
\end{table}

\paragraph{Random retrieval baseline.} %
Our target space of 79,059 words makes the random retrieval baseline:
\begin{equation}
    \text{P@}1 = \frac{1}{N} \sum_{i=1}^{N} \frac{n_i}{U} \label{equation:baseline}
\end{equation}
where $N$ represents the total number of image classes; $i$ iterates over each image class; $n_i$ denotes the number of labels for image class $i$; $U$ refers to the total number of unique aliases. From Equation~\ref{equation:baseline}, we get P@1$\approx 0.0015\%$.

\paragraph{Length-frequency alignment baseline.} The random retrieval baseline tells us how well we can align representations across the two modalities in the absence of any signal (by chance). However, the fact that we can do better than a random baseline, does not, strictly speaking, prove that our models partially converge toward any sophisticated form of modeling the world. Maybe they simply pick up on shallow characteristics shared across the two spaces. One example is frequency: frequent words may refer to frequently depicted objects. Learning what is rare {\em is} learning about the world, but more is at stake in the debate around whether LMs understand. Or consider length: word length may correlate with the structural complexity of objects (in some way), and maybe this is what drives our alignment precision? To control for such effects, we run a second baseline aligning representations from computer vision models to two-dimensional word representations, representing words by their length and frequency. We collected frequency data based on English Wikipedia using NLTK~\cite{bird2009natural} for all aliases within our target space. We use PCA and Procrustes Analysis or ridge regression~\cite{toneva2019interpreting} to map into the length-frequency space and report the best of those as a second, stronger baseline. 

\begin{figure}
    \centering
    \includegraphics[width=1\linewidth]{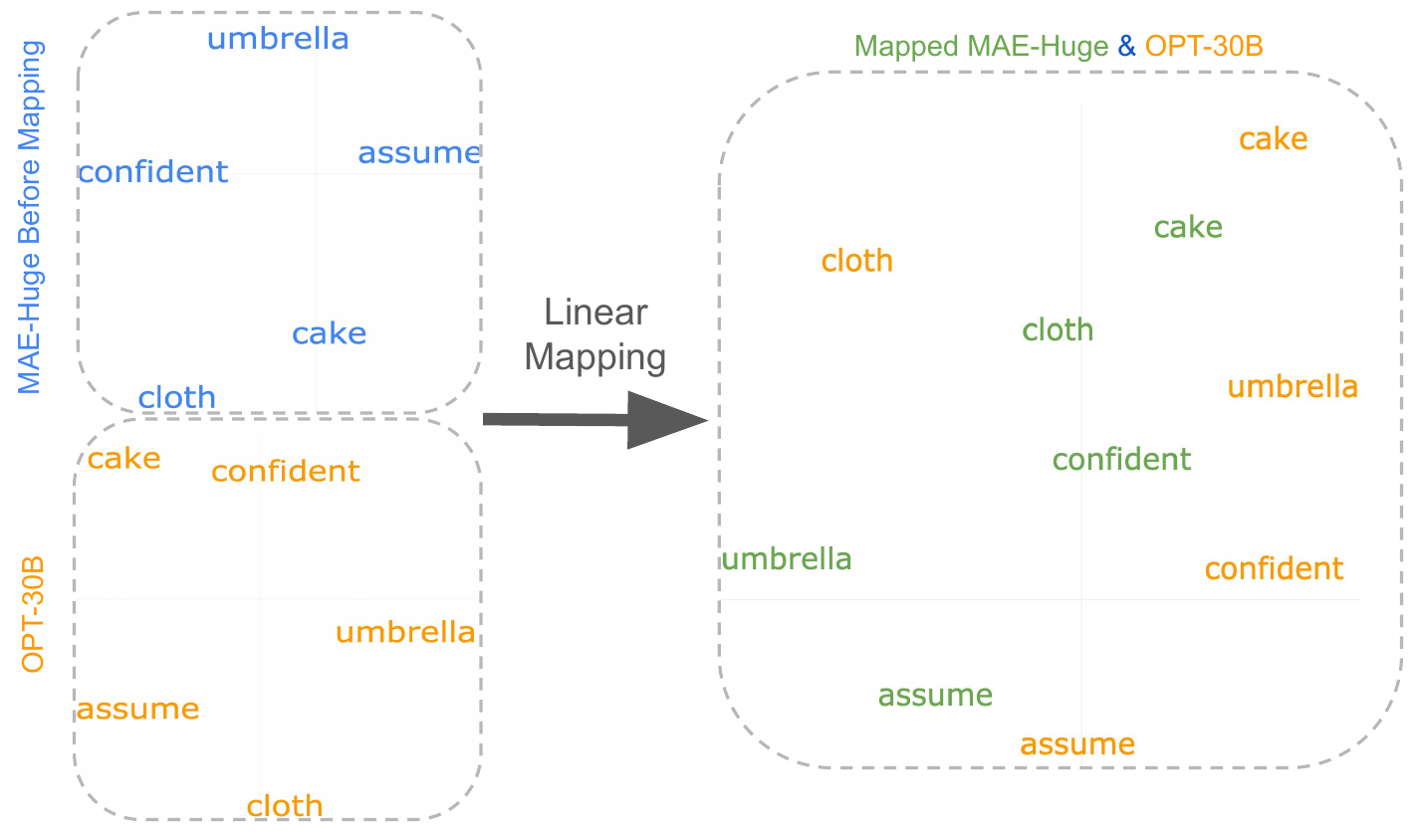}
    \caption{t-SNE plot of 5 words mapped from MAE$_{\text{Huge}}$ (\textcolor{blue}{blue}) to OPT$_{\text{30B}}$ (\textcolor{orange}{orange}) using Procrustes analysis. The \textcolor{darkgreen}{green} represent the mapped MAE$_{\text{Huge}}$ embeddings.}
    \label{fig:tsne_mapped_res}
\end{figure}

\begin{figure*}[ht]
    \centering
    \includegraphics[width=1\textwidth]{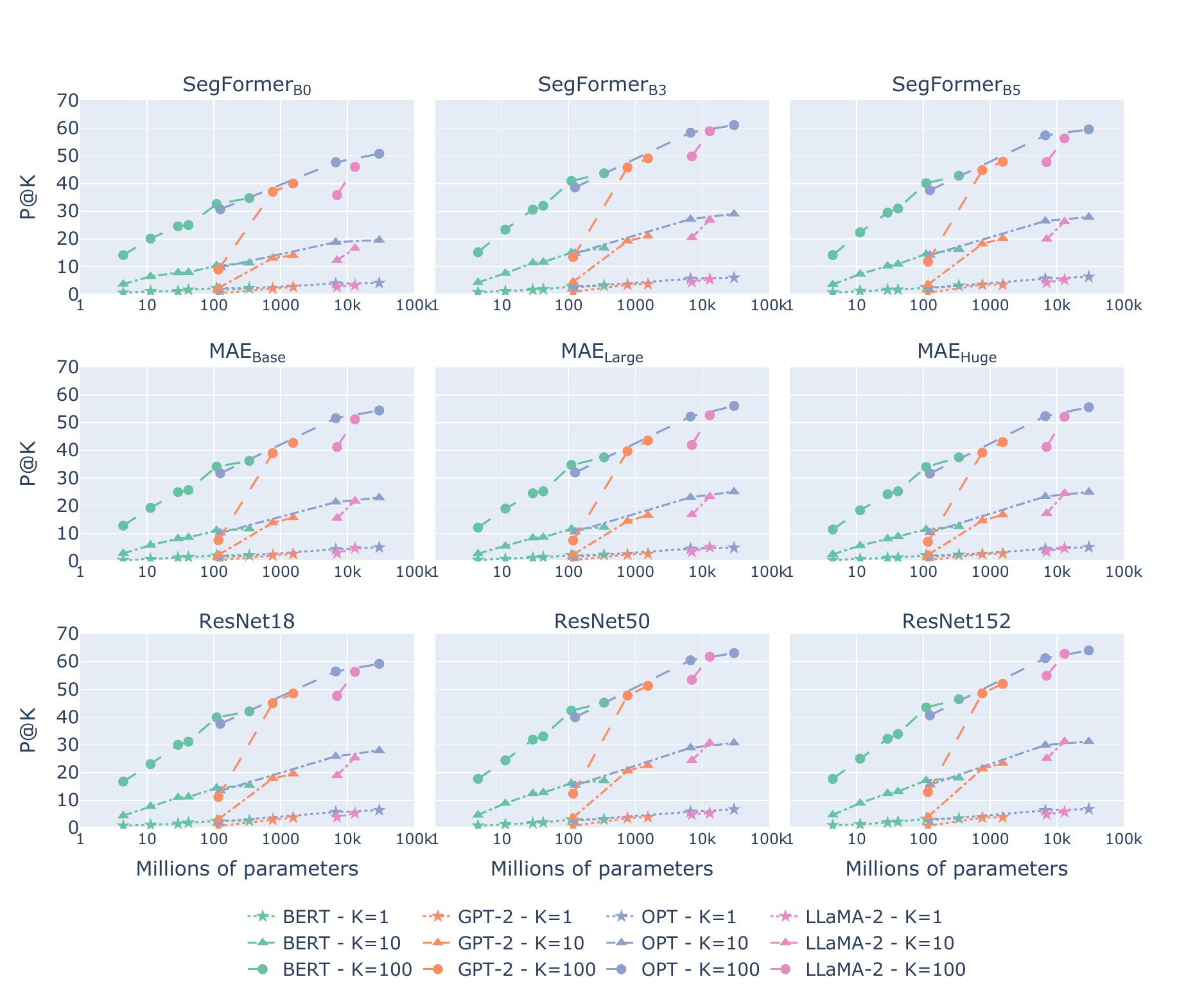}
    \caption{LMs converge toward the geometry of visual models as they grow larger on Exclude-1K set. %
    \label{fig:seg_mae_res-lms-cleaned}}
\end{figure*}

\paragraph{Non-isomorphic alignment baseline.} The former two baselines examine the possibility of aligning representations across two modalities based on chance or shallow signals. While informative, neither strictly demonstrates that a linear projection cannot effectively establish a connection between two non-isomorphic representation spaces, potentially outperforming the random or length-frequency baselines. To rigorously explore this, we disrupt the relationship between words and their corresponding representations by shuffling them. This permutation ensures that the source and target spaces become non-isomorphic. Specifically, we shuffled OPT$_{\text{30B}}$ three times at random and report the alignment results between those and original OPT$_{\text{30B}}$, we use the same Procrustes analysis for computing the alignment. Table~\ref{tab:baseline-details} presents a comparison of the three different baselines. All baselines have P@100 well below 1\%. Our mappings between VMs and LMs score much higher (up to 64\%), showing the strength of the correlation between the geometries induced by these models with respect to a conservative performance metric.

%% file: sections/05_res.tex
\section{Results}
\begin{figure*}[ht]
    \centering
    \includegraphics[width=1\textwidth]{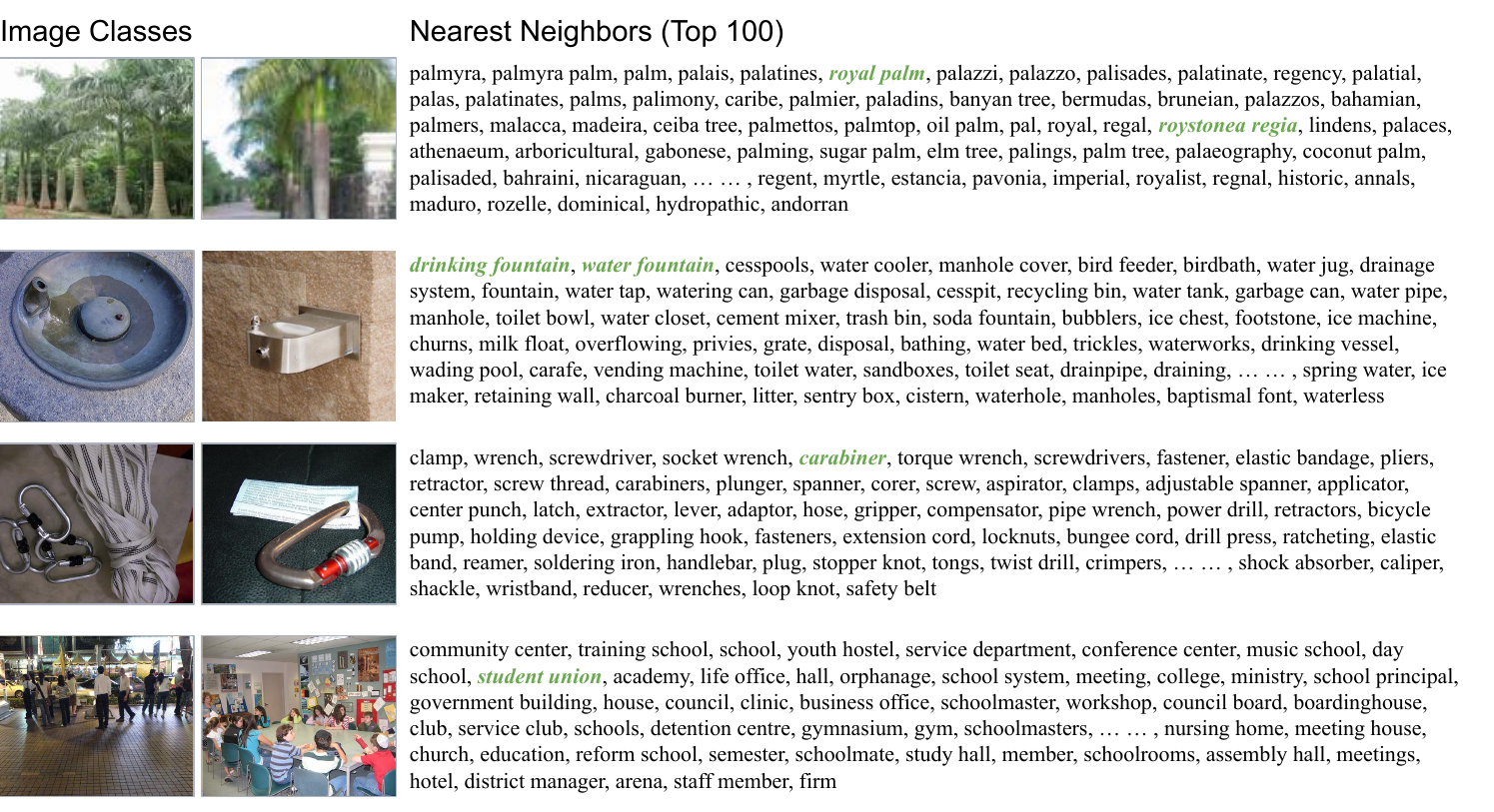}
    \caption{Examples featuring the 100 nearest neighbors in the mapping of image classes into the language representation space (from MAE$_{\text{Huge}}$ to OPT$_{\text{30B}}$). The golden labels are highlighted in \textcolor{darkgreen}{green}.}
    \label{fig:res-examples-more}
\end{figure*}
Similarities between visual and textual representations and how they are recovered through Procrustes Analysis are visualized through t-SNE in Figure~\ref{fig:tsne_mapped_res}. %
Our main results for nine VMs and all LMs are presented in Figure~\ref{fig:seg_mae_res-lms-cleaned}. The best P@100 scores are around 64\%, with baseline scores lower than 1\% (Table~\ref{tab:baseline-details}). In general, even the smallest language models outperform the baselines by orders of magnitude. 
We focus mainly on P@10 and P@100 scores because 
P@1 only allows one surface form to express a visual concept, but in reality, an artifact such as a vehicle may be denoted by many lexemes (car, automobile, SUV, etc.), each of which may have multiple inflections and derivations (car, cars, car's, etc.). %
Figure~\ref{fig:res-examples-more} shows examples where the top predictions seem `as good' as the gold standard. We find that a region of 10 neighbours corresponds roughly to grammatical forms or synonyms, and a neighbourhood of 100 word forms corresponds roughly to coarse-grained semantic classes.
Results of P@10 in Figure~\ref{fig:seg_mae_res-lms-cleaned}, show that up to one in five of all visual concepts were mapped to the correct region of the language space, with only a slight deviation from the specific surface form. Considering P@100, we see that more than two thirds of the visual concepts find a semantic match in the language space when using ResNet152 and OPT or LLaMA-2, for example. We see that ResNet models score highest overall, followed by SegFormers, while MAE models rank third. We presume that this ranking is the result, in part, of the model's training objectives: Object classification may induce a weak category-informed bias into the ResNet encoders.
In this sense, the performance of MAE models, which are fully unsupervised, presents the strongest evidence for the alignability of vision and language spaces---the signal seemingly could not come from anywhere else but the intrinsic properties of the visual world encoded by the models.

\begin{figure*}[ht]
    \centering
    \includegraphics[width=0.49\linewidth]{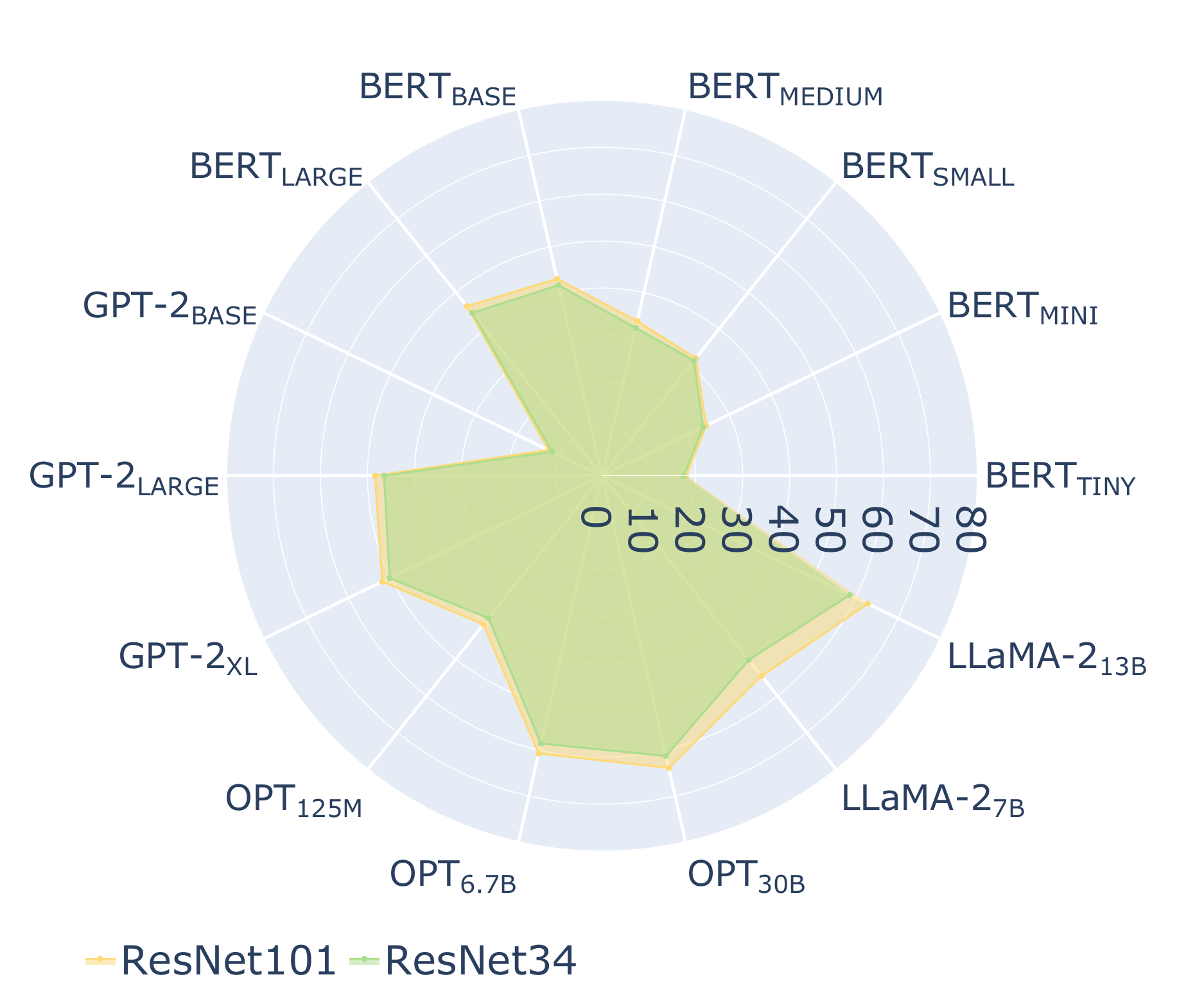}
    \includegraphics[width=0.49\linewidth]{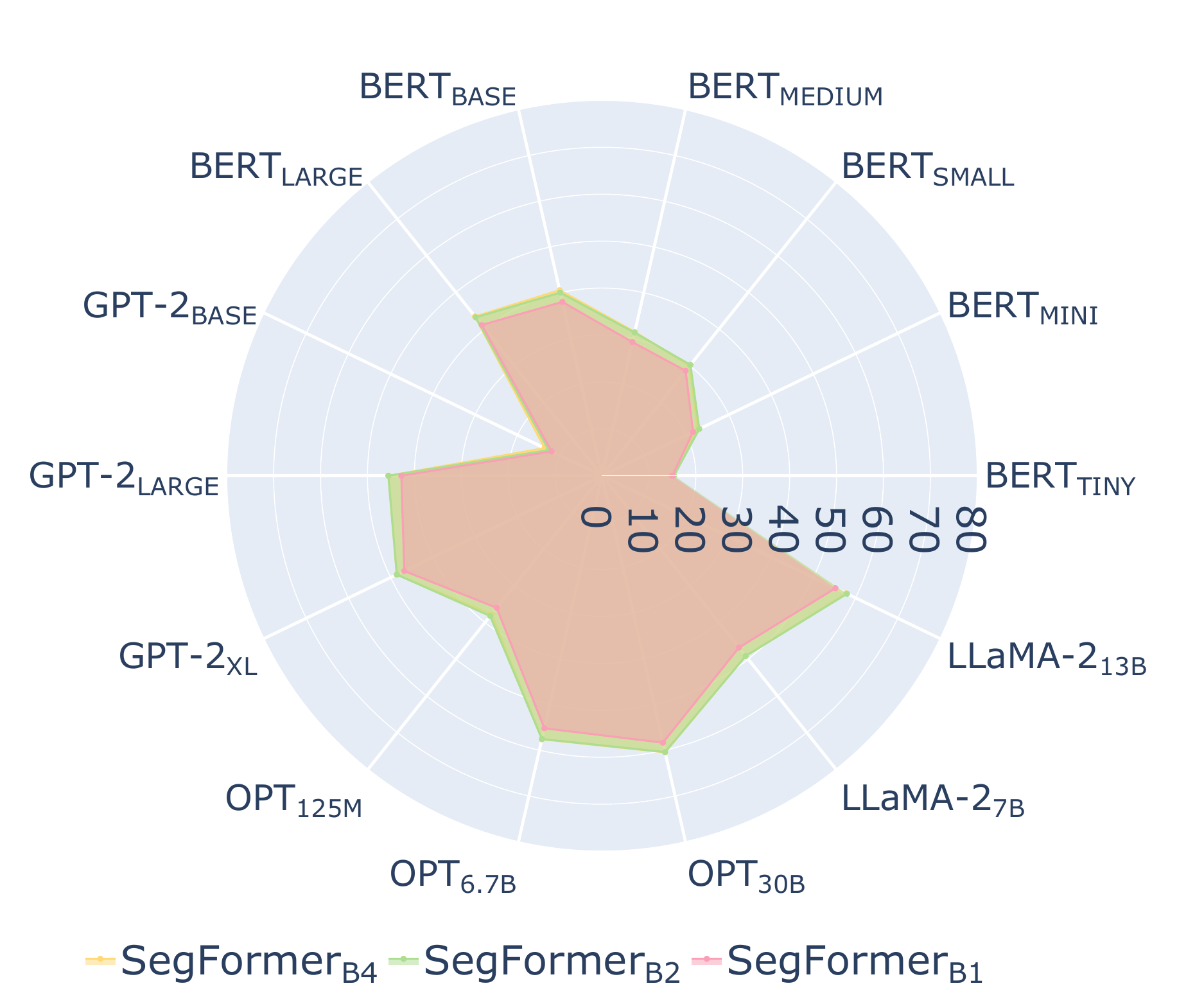}
    \caption{Illustrating the impact of scaling VMs up on Exclude-1K set. The incremental growth in P@100 for scaled-up VMs is marginal, contrasting with the more substantial increase observed when scaling up LMs in the same family.}
    \label{fig:left-models}
\end{figure*}

In Figure~\ref{fig:seg_mae_res-lms-cleaned} and \ref{fig:left-models}, we see a clear trend: as model size increases, structural similarity to VMs goes up. The correlation between VM size and similarity is slightly weaker. Specifically,
ResNet152 and OPT$_{\text{30B}}$ obtain the best results, with a P@100 of 64.1\%, i.e., 6/10 visual concepts are mapped onto a small neighborhood of 100 words---out of total set of 79,059 candidate words. Around 1/3 images are correctly mapped onto neighborhoods of 10 words, and about 1/20, onto exactly the right word. %
The scaling effect seems log-linear, with no observed saturation.%
\footnote{We also investigate the effects of incorporating text signals during vision pretraining by comparing pure vision models against selected CLIP vision encoders. The findings are unsurprising -- more details are presented in Appendix~\ref{sec:appendix-clip-results}.}

%% file: sections/06_analysis.tex
\section{Analysis}
Here, we test whether our findings extend to different parts of speech, as well as how alignment precision is influenced by factors such as dispersion, polysemy, and frequency, or by the size of the seed used for Procrustes analysis. 
For all experiments in \S~6, we use the largest model per model family for both VMs and LMs and the Exclude-1K bimodal dictionary, unless stated otherwise. %

\begin{table}
\centering
\resizebox{1\linewidth}{!}{
\begin{tabular}{l|ll|rrr} 
\toprule
  Models & Train & Test & P@1 & P@10 & P@100\\
\midrule
{MAE$_{\text{Huge}}$} & \multirow{3}{0.8cm}{Noun 1337} & \multirow{3}{0.8cm}{Adj. 157} & 2.5 & 15.9 & 50.9 \\
{SF-B5} & & & \textbf{3.2} & 19.1 & 53.5\\
{ResNet152} & & &  0.6 & \textbf{22.9} & \textbf{57.3} \\
\midrule
{MAE$_{\text{Huge}}$} & \multirow{3}{0.8cm}{Noun 1337} & \multirow{3}{0.8cm}{Verb 196} & 0.5 & 8.7 & 49.5 \\
{SF-B5} & & & \textbf{1.0} & 10.7 & 56.1\\
{ResNet152} & & &  0.0 & \textbf{13.8} & \textbf{61.2}\\
\midrule
{MAE$_{\text{Huge}}$} & \multirow{3}{0.8cm}{Mix 1337} & \multirow{3}{0.8cm}{Mix 353} & 6.7 & 45.6 & 77.8 \\
{SF-B5} & & & \textbf{6.9} & 48.8 & 80.1\\
{ResNet152} & & &  4.8 & \textbf{51.7} & \textbf{81.9}\\
\bottomrule
\end{tabular}}
\caption{\label{tab:pos_res} Evaluation of POS impact on OPT$_{\text{30B}}$ and largest VMs across various families, by showing the influence of different POS on EN-CLDI set. "Mix" denotes a combination of all POS categories.}
\end{table}

\paragraph{Part of speech.\label{analysis:pos}} ImageNet mostly contains nouns. To measure the generalization of the learned mapping to other parts of speech, we train and/or test it on adjectives, verbs, and nouns from EN-CLDI~\citep{hartmann-sogaard-2018-limitations}. %
The target space (language) in this dataset consists of 1690 concepts filtered from 79,059 concepts, which leads to a P@100 baseline below 8\%. %
We consider two settings: (1) to evaluate whether our approach is robust across parts of speech we use concepts of all POS (nouns, adjectives, and verbs) as training and evaluation data; (2) to evaluate whether the mapping learned by nouns generalizes to other POS, we use {\em only nouns} as training data and evaluate on adjectives and verbs. The results are in Table~\ref{tab:pos_res}, along with bimodal pairs counts. In the first experiment the results are strong, for instance ResNet152 and OPT$_{\text{30B}}$ lead to a P@100 of 81.9\%. Regarding the second setting, the numbers are lower but still well above our baseline, suggesting that shared concepts between VMs and LMs extend well beyond nouns.

\begin{table}
\centering
\resizebox{\linewidth}{!}{
\begin{tabular}{lll|rrrr} 
\toprule
Models & Disp. & Pairs & BERT$_{\text{Large}}$ & GPT-2$_{\text{XL}}$  & LLaMA-2$_{\text{13B}}$ & OPT$_{\text{30B}}$\\ 
\midrule
\multirow{3}{*}{SF-B5}  
& low & 703.8 & 43.2 & 47.8 & \textbf{58.7} & \textbf{60.6}\\
& med. & 744.4& 42.2 & \textbf{48.7} & 56.9 & 60.4\\
& high & 714.8& \textbf{43.3} & 47.1 & 53.3 & 57.6\\
\midrule
\multirow{3}{*}{MAE$_{\text{Huge}}$}     
& low &847.6 & \textbf{40.9} & \textbf{44.7} & \textbf{54.9} & \textbf{56.7}\\
& med. &683.8& 37.7 & 43.6 & 53.8 & 55.3\\
& high &631.6& 32.7  & 40.1 & 46.7 & 54.3\\
\midrule
\multirow{3}{*}{ResNet152}     
& low &683.0& \textbf{50.6} & \textbf{57.6} & \textbf{71.3} & \textbf{70.5}\\
& med. &739.8& 45.8 & 51.8 & 60.8 & 63.4\\
& high &740.2& 43.1 & 46.8 & 56.7 & 58.4\\
\bottomrule
\end{tabular}}

\caption{\label{tab:img_dispersion} Effect of image dispersion on mapping performance of various LMs and VMs  across different levels of image dispersion in terms of P@100 scores on the Exclude-1K set. SF=SegFormer.}
\end{table}

\begin{table*}
\centering
\resizebox{1\linewidth}{!}{
\begin{tabular}{l|lr|rrr||lr|rrr}
\toprule
Models & Polysemy & Pairs & MAE$_{\text{Huge}}$ & ResNet152 & SF-B5 & Frequency Rank & Pairs & MAE$_{\text{Huge}}$ & ResNet152 & SF-B5 \\
\midrule
& 1   & 87.0  &\textbf{ 38.3} &\textbf{ 47.3} & \textbf{42.5} & 0-10k & 96.6 & 23.0 & 25.0 & 23.7  \\ 
BERT$_{\text{Large}}$& 2-3 & 89.0  &   24.4 & 29.7 & 28.1    & 10k-100k & 700.0 & 34.3 & \textbf{44.8} & 40.1 \\
& 4+  & 106.4  & 17.4 & 23.6 & 19.2   & 100k+ & 1366.4  & \textbf{35.3} & 43.3 & \textbf{40.6} \\
\midrule

& 1   & 87.0   & \textbf{42.1} & \textbf{52.4} & \textbf{48.9} & 0-10k  &96.6 & 21.0 & 22.1 & 22.0 \\
GPT-2$_{\text{XL}}$  & 2-3 & 89.0  & 28.0 & 34.9 & 32.7   & 10k-100k & 700.0  & 39.7 & \textbf{50.4} & \textbf{45.4} \\
& 4+  & 106.4  & 15.7 & 22.8 & 21.0 & 100k+ & 1366.4 & \textbf{40.2} & 48.7 & 44.7 \\
\midrule

& 1   & 87.0 & \textbf{40.6} & \textbf{51.8} & \textbf{48.9} & 0-10k & 96.6   & 18.7 & 21.6 & 18.1 \\
LLaMA-2$_{\text{13B}}$    & 2-3 & 89.0  & 26.1 & 35.5 & 31.3 & 10k-100k & 700.0  & 40.8 & 52.3 & 46.9 \\
& 4+ & 106.4   & 14.8 & 21.3 & 18.1 & 100k+ & 1366.4 & \textbf{53.1} & \textbf{64.0} & \textbf{57.0} \\
\midrule
& 1   & 87.0 &  \textbf{47.4} & \textbf{57.2} & \textbf{55.9}  & 0-10k & 96.6 & 24.5 & 24.7 & 25.4 \\
OPT$_{\text{30B}}$    & 2-3 & 89.0      & 31.8 & 39.6 & 34.4 & 10k-100k & 700.0 & 47.2 & 56.1 & 52.3 \\
& 4+ & 106.4    & 19.9 & 25.4 & 25.2   & 100k+  & 1366.4 & \textbf{55.2} & \textbf{64.0} & \textbf{59.0} \\
\bottomrule
\end{tabular}}
\caption{\label{tab:lang_polysemy_freq} Comparison of different levels of alias polysemy and frequency on mapping performance in terms of P@100 scores on the Exclude-1K set. SF=SegFormer.}
\end{table*}

\begin{table}
\centering
\resizebox{1\linewidth}{!}{
\begin{tabular}{l|lr|rrr}
\toprule
Models & Dispersion & Pairs & {MAE$_{\text{Huge}}$} & {ResNet152} & {SF-B5} \\
\midrule
& low & 1100.6 &\textbf{44.1} & \textbf{54.5} & \textbf{51.2}\\ 
BERT$_{\text{Large}}$ & medium & 571.8 & 30.7 & 38.4 & 34.1\\
& high & 490.6 & 16.9 & 21.8 & 19.1\\
\midrule

& low & 815.2 & \textbf{46.3} & \textbf{53.9} & \textbf{50.3}\\
GPT-2$_{\text{XL}}$  & medium & 768.8 & 42.0 & 52.3 & 47.0\\
& high & 579.0 & 25.2 & 34.3 & 30.7\\
\midrule

& low & 702.8 &  \textbf{58.7} & \textbf{68.5} & \textbf{62.3}\\
LLaMA-2$_{\text{13B}}$ & medium & 707.8 & 52.9 & 65.1 & 57.3\\
& high & 752.4 & 32.3 & 42.1 & 36.8\\
\midrule

& low & 779.8 &  \textbf{60.4} & \textbf{67.6} & \textbf{63.2}\\
OPT$_{\text{30B}}$    & medium &721.2 & 55.1 & 65.1 & 59.7\\
& high &662.0 & 35.6 & 43.7 & 40.4 \\
\bottomrule
\end{tabular}}
\caption{\label{tab:lang_disp} Effect of language dispersion on mapping performance of various LMs and VMs across different levels of language dispersion in terms of P@100 scores on the Exclude-1K set. SF=SegFormer.}
\end{table}

\begin{figure}[ht]
    \centering
    \includegraphics[width=1\linewidth]{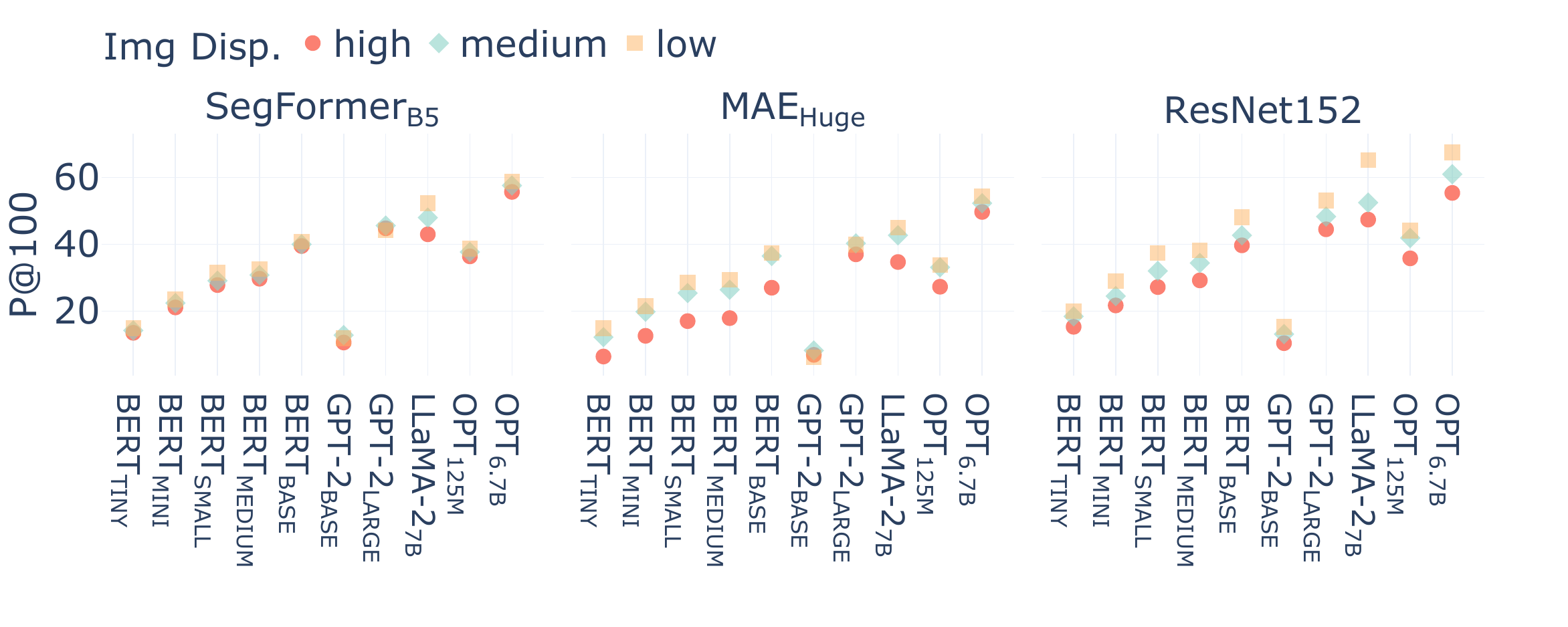}
    \includegraphics[width=1\linewidth]{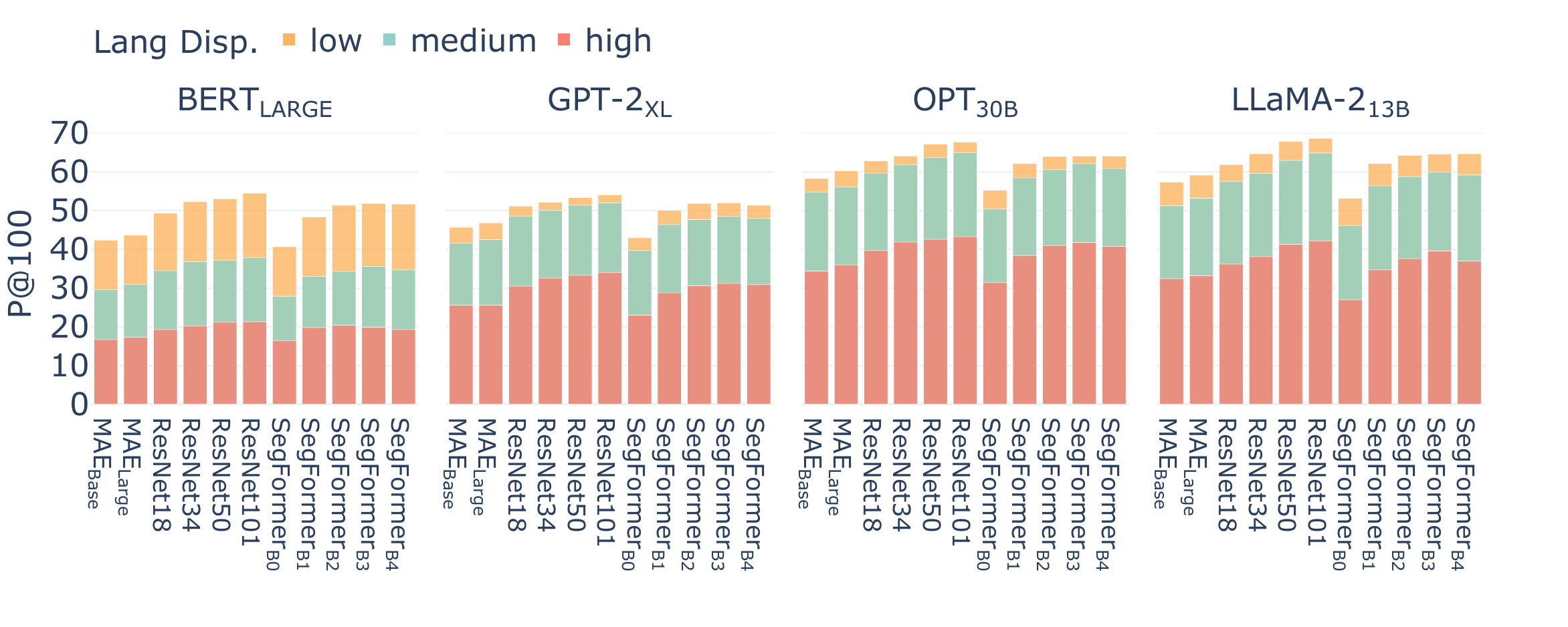}
    \includegraphics[width=1\linewidth]{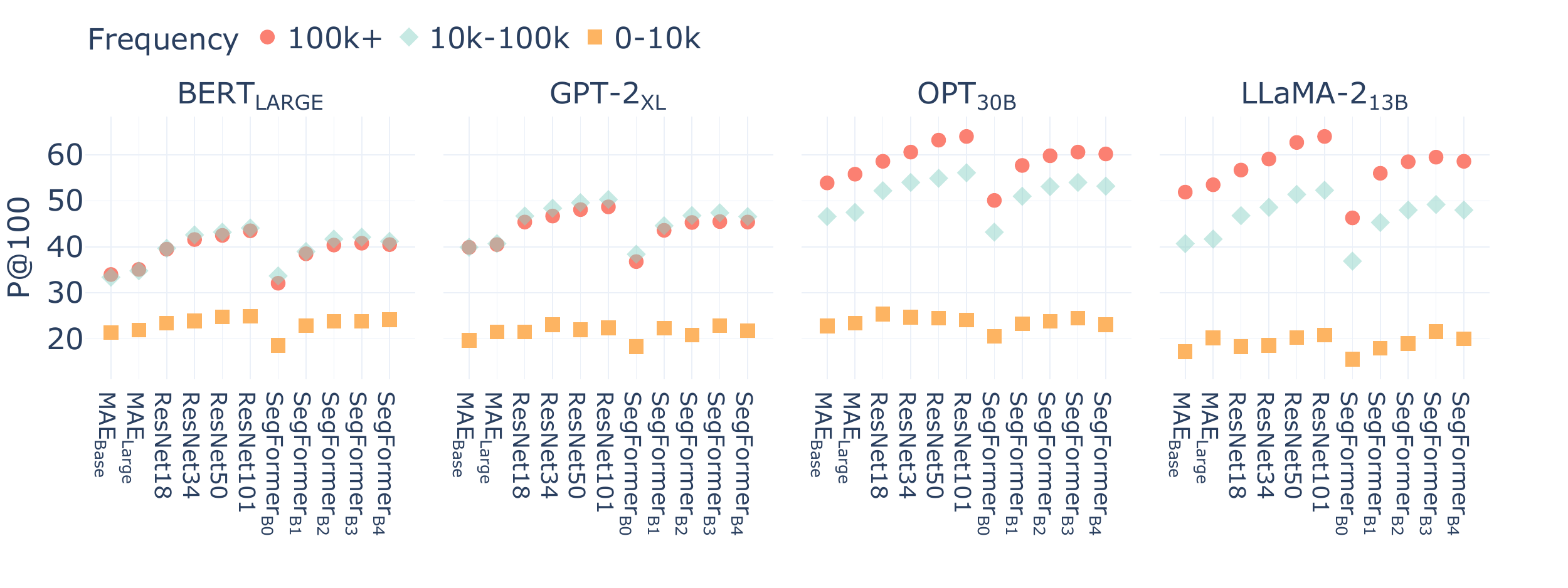}
    \includegraphics[width=1\linewidth]{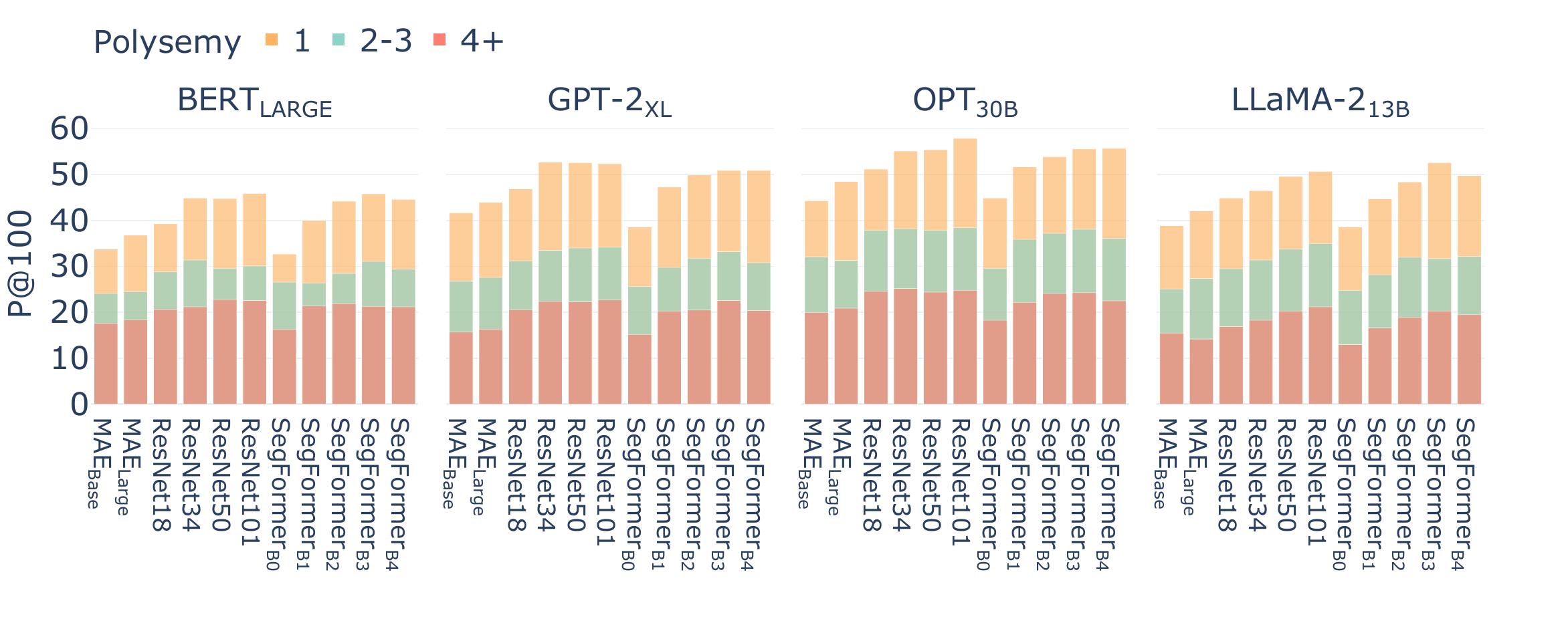}
    \caption{Performance of LMs and VMs across varied levels of (from top to bottom) image dispersion, language dispersion, frequency, and polysemy. Results are presented in terms of P@100 on the Exclude-1K dataset. Img=Image; Lang =Langauge; Disp=Dispersion.}
    \label{fig:remain_analysis}
\end{figure}

\paragraph{Image dispersion.\label{analysis:disp-poly-freq}}  %
Image dispersion is calculated by averaging the pair-wise cosine distance between all images associated with a concept~\cite{kiela-etal-2015-visual}.
We partition all concepts within the bimodal dictionary into three equally-sized distributed bins (low, medium, high) based on their dispersion. Subsequently, we classify the held-out concepts into these bins and present the results in Table~\ref{tab:img_dispersion}.
Results for ResNet152 and MAE$_{\text{Huge}}$ show concepts of lower dispersion are easier to align, while results for SegFormer-B5 are mixed.
See Figure~\ref{fig:remain_analysis} for the same consistent results observed across the remaining LMs.

\paragraph{Polysemy.} %
Words with multiple meanings may have averaged-out LM representations, %
and as such we would expect higher polysemy to cause a drop in precision. %
We obtain polysemy counts from BabelNet~\cite{10.1016/j.artint.2012.07.001} 
for the aliases in our bimodal dictionary and measure precision over non-polysemous words, words with two or three meanings, and words with four and more meanings. We ignore aliases not in BabelNet. Precision scores for these three bins are presented in Table~\ref{tab:lang_polysemy_freq}, alongside counts of the image-text pairs located in each bin. Unlike other results reported in the paper,  precision is computed separately for each alias, i.e. if a visual concept is associated with 4 aliases and only three of those appear among the nearest 100 neighbors, the P@100 for this concept will be reported as 75\%. This is necessary since different aliases for the same concept may land in different bins. The trend, as expected, is for non-polysemous aliases to yield higher precision scores, regardless of VM and LM. For ResNet152 and the largest LM in our experiments, OPT$_{\text{30B}}$, P@100 is 57.2\% for the aliases with a single meaning, but only 25.4\% for aliases with four or more meanings. 
See Figure \ref{fig:remain_analysis} for the same consistent results observed across the remaining LMs.

\paragraph{Language dispersion.} Since many of the aliases in our bimodal dictionary are {\em not} in BabelNet, e.g, multi-word expressions, we also consider `language dispersion' in Wikipedia. This is a proxy for the influence of polysemy and is measured in the same way as image dispersion. The definition and corresponding equation are in Appendix~\ref{sec:language-dispersion}.
As seen in Table~\ref{tab:lang_disp} and Figure~\ref{fig:remain_analysis}, we observe a consistent trend across all four LM families, with lower language dispersion correlating with higher alignment precision.

\paragraph{Frequency.} We proceed to investigate the influence of word frequency in our study. 
To gauge this influence, we collect and rank word frequency data from the English Wikipedia for all the unigrams and bigrams, and split them into three distinct frequency bins: the top 10,000 aliases, aliases falling within the word frequency range of 10,000 to 100,000, and aliases ranking beyond 100,000 in terms of word frequency. Then we assess precision for the aliases within our bimodal dictionary across these bins. 
The precision scores for these three bins are detailed in Table~\ref{tab:lang_polysemy_freq}, along with the corresponding counts of image-text pairs found within each bin. 
Our findings reveal a discernible trend where lower-frequency aliases consistently yield higher precision scores, for all VM and LM combinations. For instance, when utilizing ResNet152 and the most substantial LM in our experimentation, OPT$_{\text{30B}}$, P@100 reaches 64.0\% for aliases positioned beyond the 100,000 frequency mark, but decreases to 24.7\% for aliases within the top 10,000 frequency bin. 
The results of other frequency experiments for the remaining VMs and the largest LM combinations are shown in Figure~\ref{fig:remain_analysis}.

\begin{figure}[ht]
    \centering
    \includegraphics[width=1\linewidth]{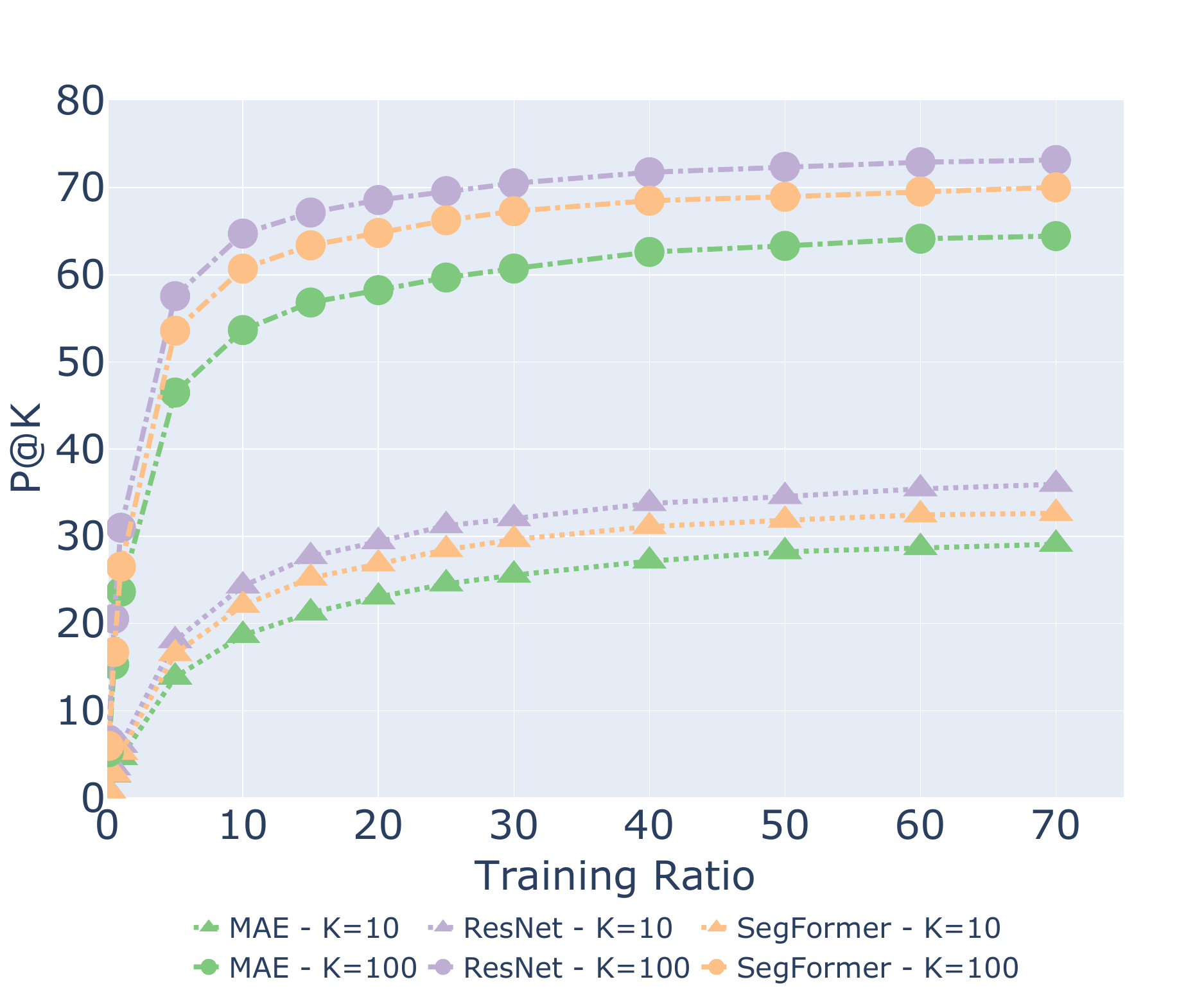}
    \caption{Effect of training data size variation on OPT$_{\text{30B}}$, for the largest vision models from three VM families.}
    \label{fig:train_ratio}
\end{figure}

\paragraph{Dictionary size.\label{analysis:train-ratio}} The training data in the bimodal dictionary---used to induce linear projections---consist of thousands of items. Here, we compare the impact of varying the size of the training data, evaluating the different mappings on the same %
5,942 concept representations within the Exclude-1K set. %
The P@100 baseline for these results is below 2\%. Mappings well above this baseline are induced with as few as 297 training pairs (5\% of the Exclude-1K dictionary). 

In sum, we have found that alignability of LMs and VMs is sensitive to image and language dispersion, polysemy, and frequency. Our alignment precision is nevertheless very high, highlighting the persistence of the structural similarities between LMs and VMs.

%% file: sections/07_discussion.tex
\section{Discussion}

Having established that language and vision models converge towards a similar geometry, we discuss the implications of this finding from various practical and theoretical perspectives within AI-related fields and beyond. %

\paragraph{Implications for the LM understanding debate.}

\citet{bender-koller-2020-climbing} has been cited for their thought experiment about an octopus listening in on a two-way dialogue between two humans stuck on deserted islands, 
but \citet{bender-koller-2020-climbing} also present the following {\em more constrained}~version of their thought experiment: 

\begin{small}
\begin{quote}{\ldots imagine training an LM (again, of any type) on English text,
again with no associated independent indications
of speaker intent. The system is also given access
to a very large collection of unlabeled photos, but
without any connection between the text and the
photos. For the text data, the training task is purely
one of predicting form. For the image data, the
training task could be anything, so long as it only
involves the images. At test time, we present the
model with inputs consisting of an utterance and
a photograph, like {\em How many dogs in the picture
are jumping?} or {\em Kim saw this picture and said
``What a cute dog'' What is cute?}}\end{quote}
\end{small}

\noindent This second thought experiment highlights the importance of relating word representations to representations of what they refer to. It also shows what their argument hinges on: If unsupervised or very weakly supervised alignment of LMs and VMs is possible, their argument fails. The question then is whether such alignment is possible? Bender and Koller deem LMs unable to `connect their utterances to the world' (or images thereof), because they assume that their representations are unrelated to representations in computer vision models. If the two representations were structurally similar, however, it would take just a simple linear mapping to make proxy inferences about the world and to establish reference. Thought experiments are, in general, fallible intuition pumps~\cite{Brendel2004-BREIPA}, and we believe our results strongly suggest that the second thought experiment of \citet{bender-koller-2020-climbing} is misleading.

\paragraph{Implications for the study of emergent properties.} The literature on large-scale, pretrained models has reported seemingly emergent properties~\cite{Søgaard2018,doi:10.1073/pnas.1907367117,Garneau2021,teehan-etal-2022-emergent,wei2022emergent}, many of which relate to induction of world knowledge. Some have attributed this to memorization, e.g.: 

\begin{small}
\begin{quote}
    It is also reasonable to assume that more parameters and more training enable better memorization
that could be helpful for tasks requiring world knowledge.~\cite{wei2022emergent}
\end{quote}\end{small} while others have speculated if this is an effect of compression dynamics~\cite{Søgaard2018,Garneau2021}. The alignability of different modalities can prove to be a very suitable test bed in the study of emergent properties relating to world knowledge. Our experiments above provide initial data points for the study of such properties. 

\paragraph{Implications for philosophy.} Our results have direct implications for two long-standing debates in philosophy: the debate around {\em strong artificial intelligence} and the so-called {\em representation wars}. Searle's original Chinese Room argument~\cite{searle80minds} was an attempt to refute artificial general intelligence, including that it was possible for a machine to {\em understand} language. Instead, Searle claims, the interlocutor in his experiment is not endowed with meaning or understanding, but mere symbol manipulation. 
Here we have showed that an interlocutor endowed only with text converges on inducing the same representational geometry as computer vision models with access to visual impressions of the world. In effect, our experiments show that some level of referential semantics (in virtue of internal models of the world) emerges from training on text alone. 

The `representation wars'~\cite{Williams2018-WILPPA-16} refers to a controversy around the role of mental representations in cognitive processes. Cognitive scientists and philosophers aligned with cognitive science often postulate discrete, mental representations to explain observed behavior. Neo-behaviourists, physicalist reductionists, and proponents of embodied cognition have argued to the contrary. To some extent, the positions have softened a bit in recent years. Many now use the term {\em representation} to mean a state of a cognitive system, i.e., neural network, that responds selectively to certain bodily and environmental conditions~\cite{Shea2018-SHERIC}. This is similar to how the term is used in machine learning and is certainly compatible with neo-behaviorism and physical reductionism. This leaves us with the problem of externalism. Proponents of embodied cognition claim that mental representations give us at best a partial story about how meaning is fixed, because meaning depends on external factors. The meaning of a proper name, for example, depends on causal factors relating the name to an initial baptism. This has also been proposed as a possible account for referential semantics in language models; see~\citet{Butlin2021-BUTSOC-2,Cappelen2021-CAPMAI,mollo2023vector,mandelkern2023language}. We believe that our experiments suggest a way to reconcile this dispute by accounting for how external factors influence our mental representations, bringing  us a bit closer to a solution to this long-standing debate.

 \paragraph{Limitations of our findings.}
Our results show that visual concepts can be mapped onto language concepts with high precision when the parameters of the mapping are learned in a supervised fashion. While this experimental setup is sufficient to uncover the structural similarities between visual and language spaces, the argument would be even stronger if we could show that the mapping can be induced in an unsupervised fashion as well, as has been done for cross-lingual embedding spaces~\cite{Conneau2018}. We experimented with algorithms for unsupervised embedding alignment~\cite{Conneau2018, artetxe2018robust, Hoshen:Wolf:18} and found that all suffer from the degenerate solution problem described in \citet{hartmann-etal-2018-unsupervised}, i.e., that no algorithm is currently available that can effectively map between modalities without supervision. We also experimented with initializing the linear transformation in unsupervised algorithms from the one learned with supervision with various amounts of noise added. We found that the unsupervised algorithms were able to recover from the offset up to a certain point, suggesting that with a sufficiently large number of random restarts, unsupervised mapping would be possible. Our experiments demonstrate linear projections can align vision and language representations for concrete noun concepts, as well as, to a lesser extent, for verbs and adjectives. Future models may enable even better linear mappings. Whether some concept subspaces are inherently (linearly) unalignable, remains an open question.

%% file: sections/08_conclusion.tex
\section{Conclusion}
In this work, we have studied the question of whether language and computer vision models learn similar representations of the world, despite being trained on independent data from independent modalities. We evaluated the structural similarity of the representations learned by these models for different sizes and architectures. We found that the geometries of these spaces are surprisingly similar, and that similarity increases with model size. These results seem to challenge the second thought experiment in \citet{bender-koller-2020-climbing}.
In our experiments, 
our baseline never goes beyond 1\% at P@100, but our linear maps exhibit P@100 scores of up to 64\%
---an inferential ability which, in our view, strongly suggests the induction of internal world models, something which many previously have deemed impossible. We have discussed various implications, but not all: In the past, researchers have speculated if image representations could act as an interlingua for cross-lingual knowledge transfer~\cite{10.5555/2283696.2283698,kiela-bottou-2014-learning,vulic-etal-2016-multi,hartmann-sogaard-2018-limitations}. Our results suggest this is viable, and that the quality of such transfer should increase log-linearly with model size.  %

%% file: sections/09_appendix.tex
\appendix

\section{Detailed Experimental Settings}
\subsection{Computational Environment}
Our primary software toolkits included HuggingFace Transformers 4.36.2~\cite{wolf-etal-2020-transformers}, PyTorch 2.1.2~\cite{NEURIPS2019_9015}. We ran our experiments on 2 NVIDIA A100s 40G. 

\subsection{Model Details}
Except for the ResNet and CLIP models, all other models used in this study are from the Huggingface Transformers (Table~\ref{tab:models-path}).
\begin{table}[ht]
\centering
\resizebox{\linewidth}{!}{
\begin{tabular}{l|c}
\toprule 
\textbf{Models} & \textbf{Links}  \\
\midrule 
BERT$_{\text{TINY}}$ & \url{https://huggingface.co/google/bert_uncased_L-2_H-128_A-2} \\
BERT$_{\text{MINI}}$ & \url{https://huggingface.co/google/bert_uncased_L-4_H-256_A-4} \\ 
BERT$_{\text{SMALL}}$ & \url{https://huggingface.co/google/bert_uncased_L-4_H-512_A-8} \\ 
BERT$_{\text{MEDIUM}}$ & \url{https://huggingface.co/google/bert_uncased_L-8_H-512_A-8} \\  
BERT$_{\text{BASE}}$ & \url{https://huggingface.co/bert-base-uncased} \\
BERT$_{\text{LARGE}}$ & \url{https://huggingface.co/bert-large-uncased} \\
\midrule
GPT-2$_{\text{BASE}}$ & \url{https://huggingface.co/openai-community/gpt2} \\
GPT-2$_{\text{LARGE}}$ & \url{https://huggingface.co/openai-community/gpt2-large}   \\
GPT-2$_{\text{XL}}$ & \url{https://huggingface.co/openai-community/gpt2-xl} \\
\midrule
OPT$_{\text{125M}}$ & \url{https://huggingface.co/facebook/opt-125m}   \\
OPT$_{\text{6.7B}}$ & \url{https://huggingface.co/facebook/opt-6.7b} \\
OPT$_{\text{30B}}$ & \url{https://huggingface.co/facebook/opt-30b} \\
\midrule
LLaMA-2$_{\text{7B}}$ & \url{https://huggingface.co/meta-llama/Llama-2-7b}   \\
LLaMA-2$_{\text{13B}}$ & \url{https://huggingface.co/meta-llama/Llama-2-13b} \\
\midrule
SegFormer-B0 & \url{https://huggingface.co/nvidia/segformer-b0-finetuned-ade-512-512}  \\
SegFormer-B1 & \url{https://huggingface.co/nvidia/segformer-b1-finetuned-ade-512-512}  \\
SegFormer-B2 & \url{https://huggingface.co/nvidia/segformer-b2-finetuned-ade-512-512}  \\
SegFormer-B3 & \url{https://huggingface.co/nvidia/segformer-b3-finetuned-ade-512-512}  \\
SegFormer-B4 & \url{https://huggingface.co/nvidia/segformer-b4-finetuned-ade-512-512}  \\
SegFormer-B5 & \url{https://huggingface.co/nvidia/segformer-b5-finetuned-ade-640-640}  \\
\midrule
MAE$_{\text{BASE}}$ & \url{https://huggingface.co/facebook/vit-mae-base}  \\
MAE$_{\text{LARGE}}$ & \url{https://huggingface.co/facebook/vit-mae-large} \\
MAE$_{\text{HUGE}}$ & \url{https://huggingface.co/facebook/vit-mae-huge}  \\
\midrule
ResNet18 & \\
ResNet34 &  \\
ResNet50 & \url{https://pypi.org/project/img2vec-pytorch/}\\
ResNet101 &  \\
ResNet152 & \\
\midrule
CLIP-RN50 & \\
CLIP-RN101 &  \\
CLIP-RN50*64 & \url{https://github.com/openai/CLIP}\\
CLIP-VIT-B-32 &  \\
CLIP-VIT-L-14 & \\
\bottomrule 
\end{tabular}}
\caption{\label{tab:models-path}
Sources of models in our experiments.}
\end{table}

\section{Dispersion Details}
\label{sec:dispersion}
\subsection{Image Dispersion}
The image dispersion $d$ of a concept alias $a$ is defined as the average pairwise cosine distance between all the image representations ${i_{1},i_{2}...i_{n}}$ in the set of $n$ images for a given alias~\cite{kiela-etal-2015-visual}:
\begin{equation}
    d(a) = \frac{2}{n(n-1)} \sum_{k<j\leq n} 1-\frac{i_{j}\cdot i_{k}}{|i_{j}||i_{k}|} \nonumber 
\end{equation}

\subsection{Language Dispersion}
\label{sec:language-dispersion}
The language dispersion $d$ of a concept alias $a$ is defined as the average pairwise cosine distance between all the corresponding word representations ${w_{1},w_{2}...w_{n}}$ in the set of $n$ sentences for a given alias:
\begin{equation}
    d(a) = \frac{2}{n(n-1)} \sum_{k<j\leq n} 1-\frac{w_{j}\cdot w_{k}}{|w_{j}||w_{k}|} \nonumber 
\end{equation}

\section{More Results}
\label{sec:more-results} 
\paragraph{Cumulative percentage of variance explained.}
In Table~\ref{tab:pca-var-ratio}, we present the cumulative percentage of variance explained by each selected component after PCA.
\begin{table}
\centering
\resizebox{1\linewidth}{!}{
\label{tab:explained-variance}
\begin{tabular}{lcccccccc}
\toprule
\multirow{2}{*}{Model} & \multicolumn{7}{c}{Explained Variance Ratio (Sum)} \\
\cmidrule(lr){2-8}
 & 256 & 512 & 768 & 1024 & 1280 & 2048 & Max \\
\midrule
MAE$_{\text{Huge}}$ & 0.9735 & 0.9922 & 0.9975 & 0.9994 & 1.0000 & - & 1.0000 \\
ResNet152 & 0.9795 & 0.9942 & 0.9974 & 0.9987 & 0.9993 & 1.0000 & 1.0000 \\
SegFormer-B5 & 0.9685 & 1.0000 & - & - & - & - & 1.0000 \\
LLaMA-2$_{\text{13B}}$ & 0.5708 & 0.6662 & 0.7277 & 0.7725 & 0.8077 & 0.8814 & 1.0000 \\
OPT$_{\text{30B}}$ & 0.4926 & 0.6002 & 0.6664 & 0.7164 & 0.7554 & 0.8360 & 1.0000 \\
\bottomrule
\end{tabular}}
\caption{\label{tab:pca-var-ratio} The cumulative of explained variance ratios for different models and sizes.}
\end{table}

\begin{table}
\centering
\resizebox{0.95\linewidth}{!}{
\begin{tabular}{l|ll|rrr} 
\toprule
  Models & Train & Test & P@1 & P@10 & P@100\\
\midrule
CLIP-ViT-L & \multirow{2}{0.8cm}{Noun 1337} & \multirow{2}{0.8cm}{Adj. 157} & \textbf{12.7} & \textbf{52.2} & \textbf{85.4} \\
CLIP-RN50x64 & & & 7.0 & 45.2 & 84.7 \\
\midrule
CLIP-ViT-L & \multirow{2}{0.8cm}{Noun 1337} & \multirow{2}{0.8cm}{Verb. 196} & \textbf{12.2} & \textbf{55.1} & \textbf{93.9} \\
CLIP-RN50x64 & & & 9.2 & 46.4 & 89.3 \\
\midrule
CLIP-ViT-L & \multirow{2}{0.8cm}{Mix 1337} & \multirow{2}{0.8cm}{Mix. 353} & \textbf{39.1} & \textbf{81.0} & \textbf{94.1} \\
CLIP-RN50x64 & & & 33.7 & 79.9 & 93.8\\
\bottomrule
\end{tabular}}
\caption{\label{tab:cldi_pos_res} Evaluation of POS impact on OPT$_{\text{30B}}$ and different CLIP models using EN-CLDI set. "Mix" denotes a combination of all POS categories.}
\end{table}

\begin{figure}
    \centering    \includegraphics[width=0.95\linewidth]{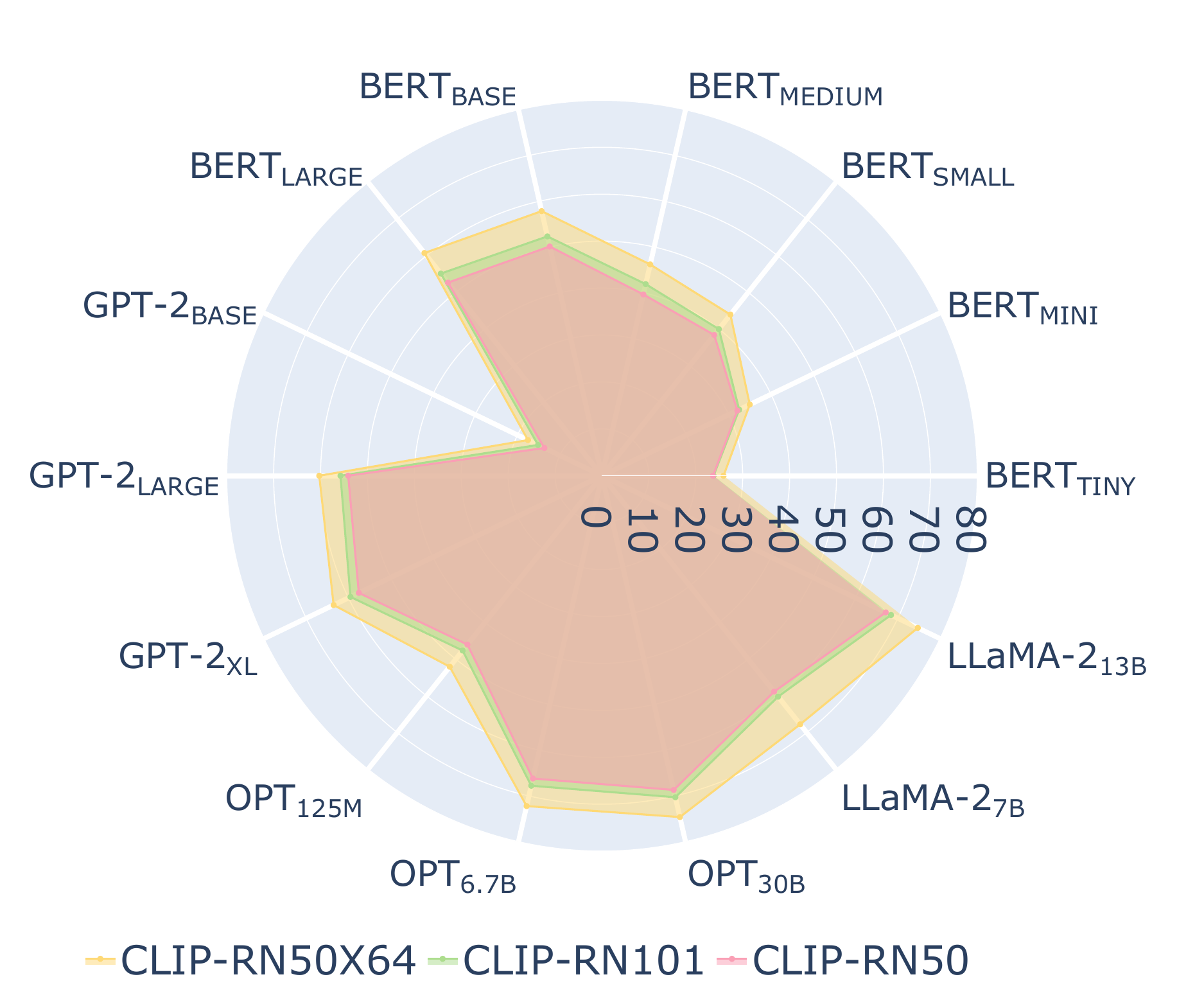}
    \includegraphics[width=0.95\linewidth]{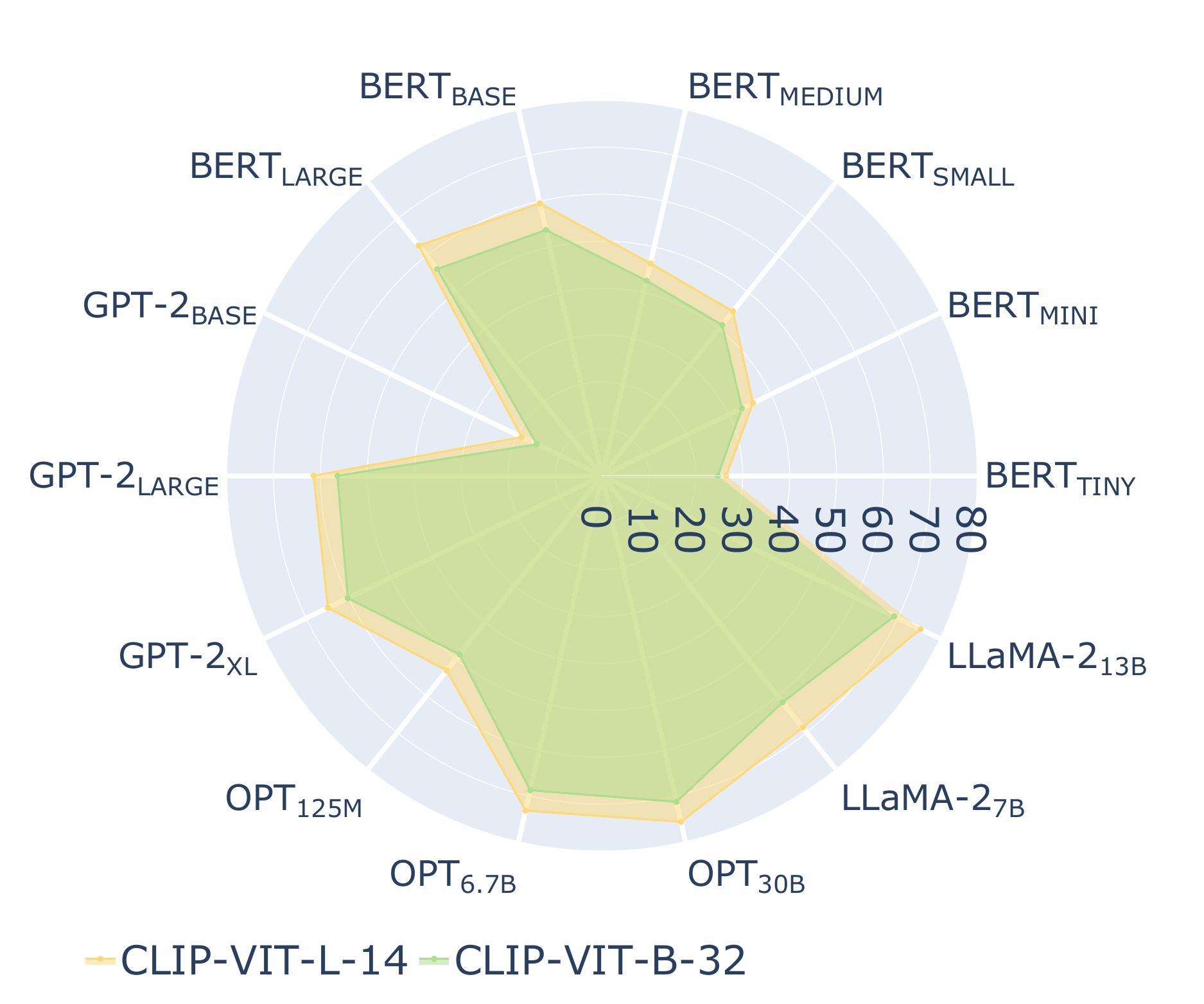}     
        \caption{Illustrating the impact of scaling CLIP models up on Exclude-1K set. The incremental growth in P@100 for scaled-up CLIP models is marginal, contrasting with the more substantial increase observed when scaling up LMs in the same family.}
    \label{fig:clip-models-p100}
\end{figure}

\begin{figure*}
    \centering
    \includegraphics[width=1\linewidth]{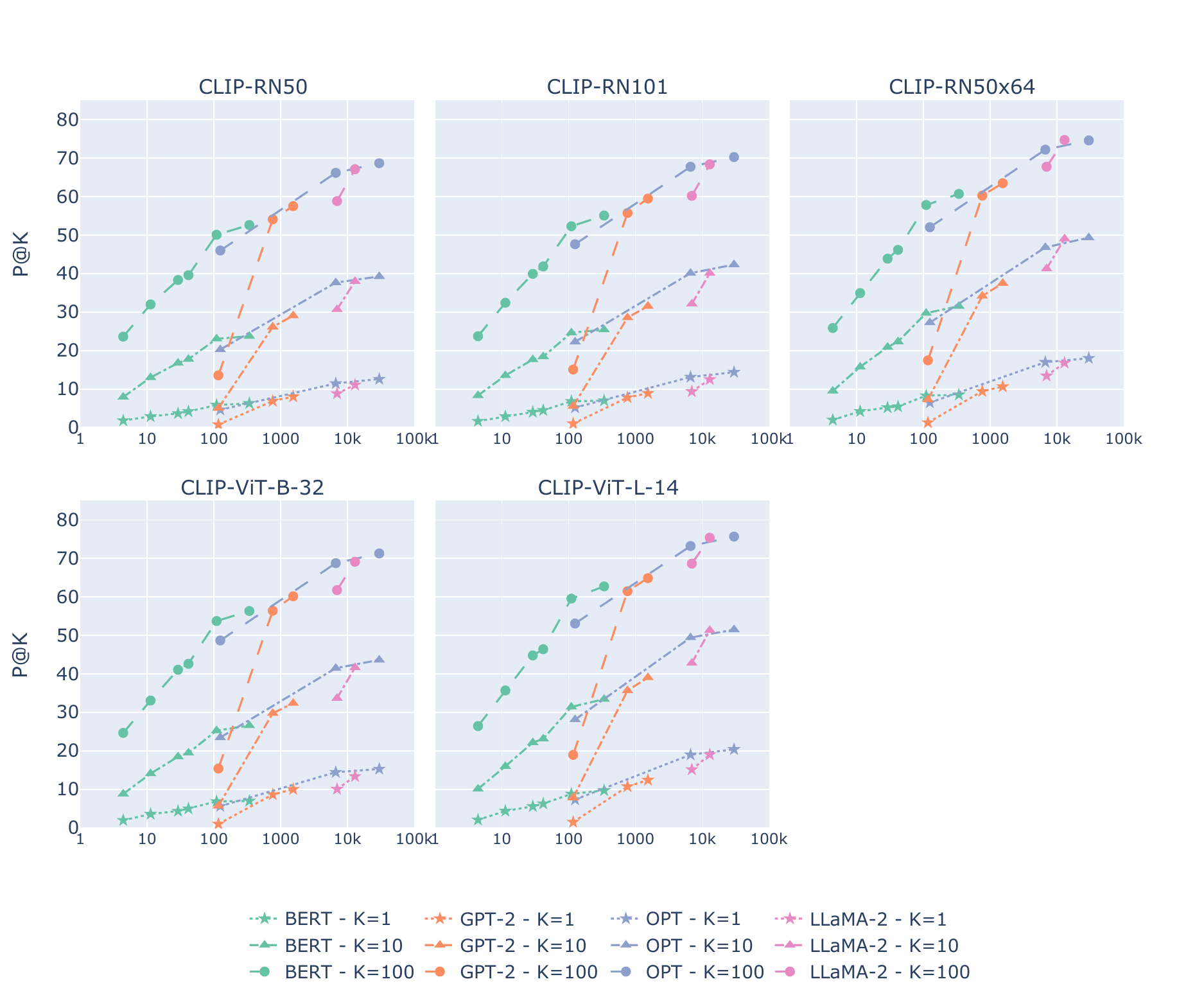}
    \caption{LMs converge toward the geometry of CLIP models as they grow larger on Exclude-1K set.}
    \label{fig:clip_all_results}
\end{figure*}

\paragraph{CLIP Results.} \label{sec:appendix-clip-results} We also investigate the effects of incorporating text signals during vision pre-training by comparing pure vision models against selected CLIP~\cite{radford2021learning} vision encoders (ResNet50, ResNet101, ResNet50x60, ViT-Base-Patch32, and ViT-Large-Patch14). The results align with our expectations, indicating that the CLIP vision encoders exhibit better alignment with LMs. The findings also support our previous observation that larger LMs tend to demonstrate better alignment. However, it would be unfair to directly compare the results from CLIP with pure vision models, as the pretraining datasets they utilize differ significantly in scale and scope. Detailed results are presented in Figure~\ref{fig:clip-models-p100} and Figure~\ref{fig:clip_all_results}.

\paragraph{POS impact on CLIP and OPT.}
In Table~\ref{tab:cldi_pos_res}, we report the POS impact on OPT$_{\text{30B}}$ and two best CLIP vision encoders in our experiments.

